\documentclass[conf]{new-aiaa}
\usepackage[utf8]{inputenc}

\usepackage{amsmath}
\usepackage[version=4]{mhchem}
\usepackage{siunitx}
\usepackage{longtable}
\usepackage{booktabs}
\usepackage[numbers]{natbib}
\usepackage{url}
\usepackage{xcolor}
\usepackage{graphicx}
\usepackage{fancyhdr}
\usepackage{multirow}
\usepackage{subcaption}
\usepackage{placeins} 
\usepackage{float}
\usepackage{tabularx}
\usepackage{indentfirst}
\usepackage{lastpage}

\begin{document}
\begin{center}
SpaceOps-2025, ID \#605\\
\vspace{1em} 
\noindent \textbf{Toward Onboard AI-Enabled Solutions to Space Object Detection for Space Sustainability}

\vspace{1em} 
\textbf{Wenxuan Zhang$^{a}$ and Peng Hu$^{ab*}$}

\vspace{1em}

\end{center}
\begin{raggedright}
$^{a}$ Faculty of Mathematics, University of Waterloo, Canada, v39zhang@uwaterloo.ca\\
$^{b}$ Department of Electrical and Computer Engineering, University of Manitoba, Canada, Peng.Hu@umanitoba.ca\\
$^{*}$ Corresponding Author\\

\end{raggedright}

\date{January 2025}
\addtolength{\topmargin}{-10.36003pt}
\pagenumbering{arabic}
\pagestyle{fancy}
\fancyhf{}
\fancyhead[C]{18th International Conference on Space Operations, Montreal, Canada, 26 - 30 May 2025. \\
Copyright ©2025 by the Canadian Space Agency (CSA) on behalf of SpaceOps. All rights reserved
}
\fancyfoot[L]{SpaceOps-2025, ID \#605}
\fancyfoot[R]{Page \thepage\ of \pageref*{LastPage}}
\renewcommand{\headrulewidth}{0pt}

\section*{Abstract}
The rapid expansion of advanced low-Earth orbit (LEO) satellites in large constellations is positioning space assets as key to the future, enabling global internet access and relay systems for deep space missions. A solution to the challenge is effective space object detection (SOD) for collision assessment and avoidance. In SOD, an LEO satellite must detect other satellites and objects with high precision and minimal delay. This paper investigates the feasibility and effectiveness of employing vision sensors for SOD tasks based on deep learning (DL) models. It introduces models based on the Squeeze-and-Excitation (SE) layer, Vision Transformer (ViT), and the Generalized Efficient Layer Aggregation Network (GELAN) and evaluates their performance under SOD scenarios. Experimental results show that the proposed models achieve mean average precision at intersection over union threshold 0.5 (mAP50) scores of up to 0.751 and mean average precision averaged over intersection over union thresholds from 0.5 to 0.95 (mAP50:95) scores of up to 0.280. Compared to the baseline GELAN-t model, the proposed GELAN-ViT-SE model increases the average mAP50 from 0.721 to 0.751, improves the mAP50:95 from 0.266 to 0.274, reduces giga floating point operations (GFLOPs) from 7.3 to 5.6, and lowers peak power consumption from 2080.7 mW to 2028.7 mW by 2.5\%.

\noindent\textbf{Keywords: } Space Object Detection, Onboard AI, Low-Earth Orbit
\makeatletter
\renewcommand\section{\@startsection {section}{1}{\z@}%
{-2.0ex \@plus -0.5ex \@minus -.2ex}%
{1.0ex \@plus.2ex}%
{\normalfont\normalsize\bfseries\raggedright}}%
\renewcommand{\thesection}{\arabic{section}}
\renewcommand{\thesubsection}{\thesection.\arabic{subsection}}

\titleformat{\subsection}
{\normalsize\itshape\singlespacing\raggedright}{\thesubsection.\space}{0pt}{#1}[]
\makeatother

\section*{Abbreviations}
\begin{tabular}{ll}
CNN & Convolutional Neural Network \\
CV & Computer Vision \\
DL & Deep Learning \\
GELAN & Generalized Efficient Layer Aggregation Network \\
LEO & Low-Earth Orbit \\
PGI & Programmable Gradient Information \\
SE & Squeeze-and-Excitation \\
SOD & Space Object Detection \\
ViT & Vision Transformer \\
YOLO & You Only Look Once \\
FOV  & Field of View \\
mAP & Mean Average Precision \\
IoU & Intersection over Union\\
mAP50 & mAP at IoU threshold 0.5 \\
mAP50:95 & mAP averaged over IoU thresholds from 0.5 to 0.95 \\
GFLOPs & Giga Floating Point Operations \\
\end{tabular}

\section{Introduction}
As advanced low-Earth orbit (LEO) satellite constellations rapidly expand, they enable global internet access and serve as relay systems for deep-space missions, making space assets increasingly important. However, ensuring the safety and sustainability of thousands of LEO satellites is a challenge. For example, the increasing collision risk between LEO satellites and other space objects can generate substantial amounts of debris of various sizes, threatening both the safe operation of spacecraft and the space environment. These risks are significant and are expected to intensify as more satellites are deployed.

Given the limited hardware capacity and energy constraints of LEO satellites, the space object detection (SOD) process needs to be highly resource-efficient. Traditional sensing technologies, such as LiDAR and ground-based radar, do not offer a high-precision, low-power, and low-latency solution required for effective SOD tasks. In this context, onboard vision sensing presents a promising alternative with its low-power, cost-effective advantages.

Early computer vision (CV) approaches relied on hand-crafted feature extraction techniques such as edge detection \cite{4767851} and optical flow tracking \cite{Lucas1981AnII}. Although these methods demonstrated reasonable success in object tracking and recognition, they often struggled in dynamic and complex environments. The advent of neural networks, particularly Convolutional Neural Networks (CNNs), has significantly advanced the field, allowing automated feature extraction and learning-based detection models to surpass traditional CV approaches in both precision and robustness. 

Modern object detection pipelines process an image to generate bounding boxes, class labels, and confidence scores for detected objects by extracting hierarchical features such as edges, textures, and object shapes. The You Only Look Once (YOLO) series, for instance, provides real-time detection suitable for onboard SOD tasks. Recent surveys in remote sensing have highlighted that deep learning-based methods have revolutionized object detection in earth observation tasks, showing strong performance across diverse data modalities such as optical, SAR, and DSM imagery \cite{rs16020327}. However, CNN-based models often struggle with small object detection due to their limited contextual awareness. In contrast, Vision Transformers (ViTs) \cite{Dosovitskiy2020AnII} capture long-range dependencies, improving small object detection at the cost of higher computational demand.

To address these challenges, recent research has introduced hybrid models that combine CNN and ViT architectures. In our previous work, we proposed two such models, GELAN-ViT and GELAN-RepViT \cite{zhang2024sensing}. In this paper, we further enhance these models by integrating Squeeze-and-Excitation (SE) \cite{Liu2021AnI2} blocks to improve channel-wise feature recalibration. SE blocks dynamically recalibrate channel-wise features to emphasize informative features while suppressing less relevant ones, improving detection performance without significantly increasing computational overhead. 

Our contributions can be summarized as follows:
\begin{itemize}
    \item We introduce SE-enhanced hybrid models - GELAN-ViT-SE, GELAN-RepViT-SE, and GELAN-SE - to improve feature selection.
    \item We develop the SODv2 dataset using the Unity engine, simulating a realistic solar system environment with dynamically varying satellite positions and occlusions.
    \item We perform a comprehensive performance analysis to compare our proposed models with state-of-the-art detection frameworks using metrics such as Giga Floating Point Operations (GFLOPs), parameter count, power consumption, inference time, and mean average precision (mAP) calculated at different intersection over union (IoU) thresholds, including mAP at IoU 0.5 (mAP50) and the mAP averaged over IoU thresholds from 0.5 to 0.95 (mAP50:95).
\end{itemize}

Section 2 begins with related work, followed by a theoretical overview and model architecture in Section 3. Section 4 discusses the generation of the SODv2 dataset, Section 5 presents our experimental results, and Section 6 concludes the paper.

\section{Related Work}
Recent advances in DL have significantly improved object detection, leading to a wide variety of neural network models tailored to different application needs. Most state-of-the-art models in this domain rely on CNNs, which are designed with built-in assumptions, such as spatial locality and translation invariance, that help them efficiently capture and learn visual features.

\subsection{CNN-based Detectors}

CNN-based detectors are generally classified into one-stage and two-stage approaches. Two-stage detectors, such as R-CNN \cite{girshick2014rich}, operate by first identifying region proposals that may contain objects and then refining these proposals through a secondary classification and regression step. While these models achieve high accuracy, their two-step nature results in higher computational costs and slower inference speeds, making them less suitable for real-time applications. On the other hand, one-stage detectors like YOLO \cite{7780460} bypass the proposal stage and perform object classification and localization in a single pass, improving the inference speed. 

The YOLO series \cite{7780460} has become one of the state-of-the-art object detection frameworks. By reformulating detection as a single regression problem from image pixels to bounding boxes and class probabilities, YOLO models focus on achieving a balance between detection speed and accuracy, making them well-suited for real-time applications. YOLOv9, introduced by \cite{wang2024yolov9learningwantlearn}, utilizes the Generalized Efficient Layer Aggregation Network (GELAN) and Programmable Gradient Information (PGI) to improve detection performance while maintaining low computational complexity.

Despite these advancements, small object detection remains a challenging task for traditional CNN-based models, as noted by \cite{Leng2019An}. While CNNs emphasize local features, which helps capture fine details in larger objects \cite{10.1007/978-3-031-25069-9_45}, this localized focus can become a limitation when objects span only a few pixels. Additionally, their focus on local relationships might hinder the model’s ability to incorporate broader contextual information, which is often essential for understanding complex scenes and accurately detecting distant objects.

\subsection{ViT-based Detectors}

ViTs \cite{Dosovitskiy2020AnII} offer a promising alternative by using self-attention mechanisms to capture long-range dependencies, allowing them to outperform CNN-based models in small object detection \cite{rekavandi2023transformerssmallobjectdetection}. However, these improvements come at the cost of increased computational complexity and higher data requirements. The self-attention mechanism leads to increased computational complexity, as mentioned in \cite{Dosovitskiy2020AnII}, which can be challenging for real-time deployments in a resource-constrained environment. Additionally, due to the lack of inductive biases, ViTs typically require larger training datasets to achieve performance levels comparable to CNNs, and are often more sensitive to hyperparameter tuning, as demonstrated in \cite{ReLU_ViT}. As a result, while ViTs offer improved global feature extraction, their practicality in certain environments may be limited by these computational and data requirements.

\subsection{Effect of SE block}

Given the challenges associated with ViTs, CNN-based architectures continue to evolve with techniques that enhance feature selection and representation learning. For example, SE blocks are introduced by Liu et al. \cite{Liu2021AnI2} to improve feature selection by dynamically recalibrating channel-wise responses. While SE blocks have been successfully applied in various vision tasks, including medical image segmentation \cite{Liu2021AnI2}, our approach leverages their strengths within GELAN-based hybrid models to enhance detection accuracy while maintaining computational efficiency. 

In the following section, we provide an in-depth discussion of our system model, explaining the effect of the SE layer and the detailed architecture of our model.

\section{The Proposed Model}
\subsection{Overview}
We propose three models: GELAN-ViT-SE, GELAN-RepViT-SE, and GELAN-SE. GELAN-ViT-SE and GELAN-RepViT-SE extend GELAN-ViT and GELAN-RepViT from our previous work \cite{zhang2024sensing} by incorporating SE blocks to improve feature selection. Following the principle of information bottleneck, which posits that neural networks have a finite capacity to process and transmit information \cite{ShwartzZiv2017OpeningTB}, these models separate feature extraction into local and global pathways. When local features (\(f_l\)) extracted by CNN and global features (\(f_g\)) extracted by ViT compete for the finite capacity of the network, resource contention may occur, leading to degraded performance in feature extraction \cite{zhang2024sensing}. Moreover, integrating ViT and CNN into a unified pipeline may cause the ViT to interfere with the CNN's inherent inductive biases, thereby reducing the efficiency of the learning process. To mitigate this, GELAN-ViT and GELAN-RepViT assign dedicated processing capacity to each branch, where a CNN pathway extracts localized spatial features and a ViT pathway captures global dependencies.

Building upon this architecture, we further improve feature extraction in the CNN pathway by integrating SE blocks. In traditional CNN-based models, uniform treatment of all channels may hinder the model's ability to prioritize informative features over less relevant ones. To improve this, we integrate SE blocks, which dynamically recalibrate channel-wise attention, enhancing the influence of critical features while reducing the influence of those that are less informative \cite{8578843}. Furthermore, we introduce GELAN-SE, a CNN-based variant derived from the GELAN-s-reduced model \cite{zhang2024sensing} with SE integration.  While both GELAN-ViT and GELAN-RepViT are built on top of GELAN-s-reduced, adding SE to the baseline GELAN-s-reduced allows for a direct comparison and helps to directly assess the impact of ViT integration on SE-enhanced models.

\subsection{SE for Efficient Capacity Allocation}

Traditional CNN architectures process feature maps uniformly across all channels, limiting their ability to dynamically emphasize the most relevant features. To overcome this limitation of uniform channel processing, the SE block \cite{8578843} introduces two operations:
\begin{itemize}
    \item Squeeze: A global average pooling computes a single scalar value \(s_c \) for each channel $c$ by averaging all spatial positions (H $\times$ W).

    \item Excitation: The aggregated values $\mathbf{s} = \{s_1, s_2, \dots, s_C\}$ are first passed through fully connected layers to generate the scaling factors $\mathbf{z} = \{z_1, z_2, \dots, z_C\}$, where $z_c \in [0,1]$. Each channel is then rescaled according to:
    \begin{equation}
    \widehat{\mathbf{X}}_c = z_c \cdot \mathbf{X}_c.\label{eq:se_scaling}
    \end{equation}

\end{itemize}

Where the scaling factors $\mathbf{z}$ are calculated by passing $\mathbf{s}$ through two fully connected layers with a ReLU activation followed by a sigmoid function, ensuring values remain between 0 and 1.

Without SE, all channels are weighted equally during the feature extraction process:

\begin{equation}
C_{\mathrm{noSE}}(F_l) = \sum_{c=1}^{C} 1,\label{eq:no_se_capacity}
\end{equation}
meaning that all $C$ channels share the local capacity uniformly.

With SE, each channel $c$ is assigned a learned weight $z_c \in [0,1]$:
\begin{equation}
C_{\mathrm{recalibrated}}(F_l) = \sum_{c=1}^{C} z_c,\label{eq:se_capacity}
\end{equation}

Channels that are deemed more informative (with larger $z_c$) contribute more to the final feature representation, while less informative channels contribute less. Although the total local capacity $C(F_l)$ remains the same, the SE block optimally redistributes it among the channels, effectively addressing internal bottlenecks within the local branch.

\subsection{Architecture}

\begin{figure}
    \centering
    \includegraphics[width=\linewidth]{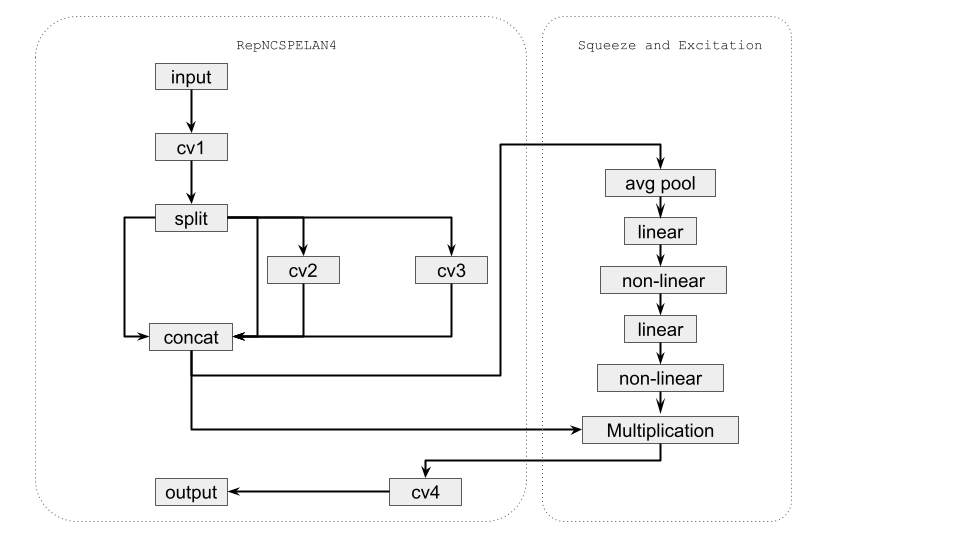}
    \caption{The architecture of the RepNCSPELAN4\_SE module. The SE layer is highlighted with a distinct background color.}
    \label{fig:SE}
\end{figure}

SE blocks are integrated into the RepNCSPELAN4 module within the last two detection heads, forming the proposed RepNCSPELAN4\_SE structure. The detailed structure of RepNCSPELAN4\_SE is shown in Fig. \ref{fig:SE}, with the steps outlined below:

\begin{enumerate}
    \item The input feature map \( x \) is first processed by an initial convolutional layer \( cv1 \), resulting in the output \( x' \).
    \item \( x' \) is then divided into two equal parts, denoted \( x_0 \) and \( x_1 \).
    \item Each branch is processed by separate convolutional layers, \( cv2 \) and \( cv3 \), producing \( x_0' \) and \( x_1' \), respectively.
    \item The feature maps \( x_0 \), \( x_1 \), \( x_0' \), and \( x_1' \) are concatenated along the channel dimension to form an intermediate feature representation \( y \).
    \item The SE block then recalibrates \( y \) by adjusting the channel-wise attention as follows:
    \begin{enumerate}
            \item The global average pooling computes a channel descriptor \( s_c \) for each channel.
            \item The pooled features are passed through two fully connected layers with non-linear activations to generate attention weights \( z_c \).
            \item These attention weights are applied element-wise to the feature map, resulting in a recalibrated output \( y' \).
        \end{enumerate}
    \item Finally, the recalibrated feature map \( y' \) is passed through a final convolutional layer, denoted \( cv4 \), to generate the final output.
\end{enumerate}

\begin{figure}[H]
    \centering
    \begin{subfigure}{0.7\linewidth}
        \centering
        \includegraphics[width=\linewidth]{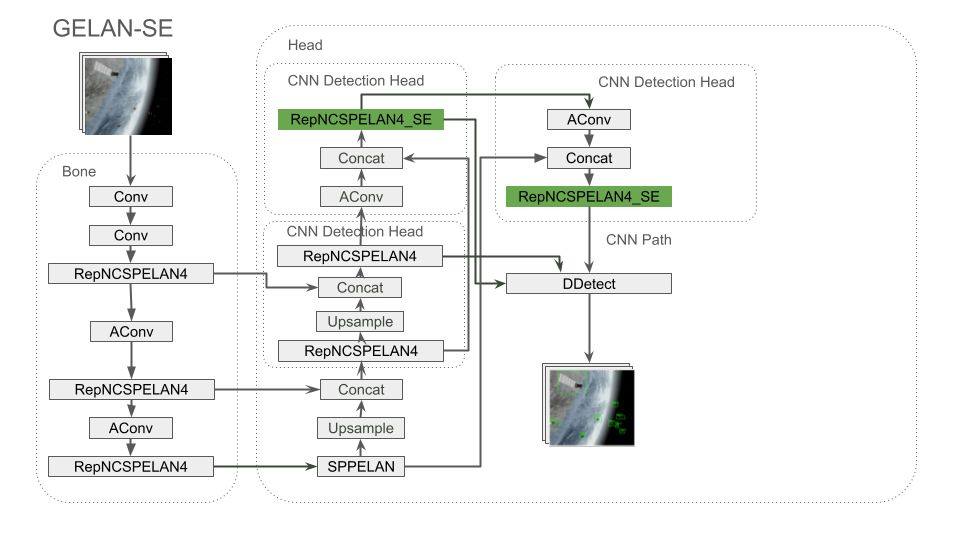}
        \caption{The architecture of the GELAN-SE model.}
        \label{fig:GELAN_SE}
    \end{subfigure}
    
    \begin{subfigure}{0.7\linewidth}
        \centering
        \includegraphics[width=\linewidth]{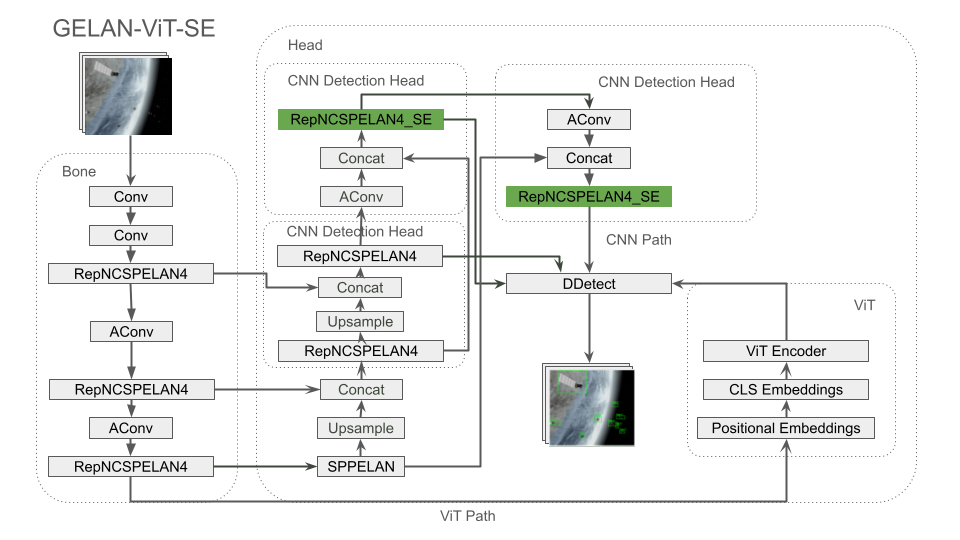}
        \caption{The architecture of the GELAN-ViT-SE model.}
        \label{fig:GELAN_ViT_SE}
    \end{subfigure}

    \begin{subfigure}{0.7\linewidth}
        \centering
        \includegraphics[width=\linewidth]{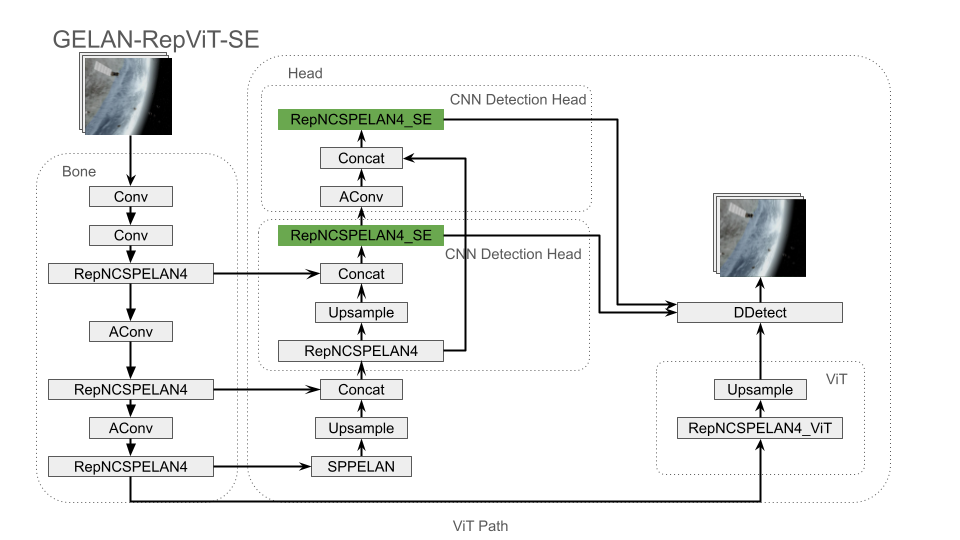}
        \caption{The architecture of the GELAN-RepViT-SE model.}
        \label{fig:GELAN_RepViT_SE}
    \end{subfigure}

    \caption{Architectures of the proposed GELAN-based models. The SE block is integrated into the RepNCSPELAN4\_SE structure.}
    \label{fig:GELAN_models}
\end{figure}

The architectures of GELAN-SE, GELAN-ViT-SE, and GELAN-RepViT-SE are illustrated in Fig. \ref{fig:GELAN_models}, with each design outlined below:

\begin{itemize}
    \item \textbf{GELAN-SE:} As illustrated in Fig. \ref{fig:GELAN_SE}, GELAN-SE is a purely CNN-based model that improves the local feature extraction through SE integration.
    
    \item \textbf{GELAN-ViT-SE:} As illustrated in Fig. \ref{fig:GELAN_ViT_SE}, GELAN-ViT-SE extends the GELAN backbone by introducing a ViT pathway parallel to the CNN head. The SE layer is integrated to refine local feature extraction within the CNN pathway.
    
    \item \textbf{GELAN-RepViT-SE:} As illustrated in Fig. \ref{fig:GELAN_RepViT_SE}, GELAN-RepViT-SE is designed for computational efficiency. It incorporates a streamlined ViT encoder denoted as RepNCSPELAN4\_ViT, within the RepNCSPELAN4 module, enabling efficient feature fusion with minimal computational overhead. The SE layer further enhances local feature extraction in the CNN pathway.

\end{itemize}

To support the evaluation of our models, we next describe the SODv2 dataset generation process.

\section{SODv2 Generation}

The SODv2 dataset is generated using Unity to simulate a realistic solar system, in which the scale, distances, and motion of celestial bodies and LEO satellites reflect real-world physics. In the simulation, LEO satellites are initially spawned at altitudes randomly chosen between 500 km and 600 km above Earth's surface, with random orbital placements. Random orbits rarely result in satellites being positioned within each other's field of view (FOV). To address this, we cluster the satellites to ensure every simulated satellite has at least one neighbor. Clustering begins with an initial set of satellites that serve as cluster centers. Around each cluster center, a random number of additional satellites (between 1 and 20) are placed within a 5 km radius. Each simulation cycle generates 1,000 satellites.

The simulation replicates an onboard satellite camera with a fixed FOV. In each frame, the camera is attached to a specific satellite, capturing the surrounding environment with a fixed 45-degree FOV. The simulation script identifies the closest satellite to the camera and adjusts the camera angle to ensure the target satellite is captured in the frame. The script then records the metadata along with the image, including the distances between the camera and all visible satellites. After an image is captured, the camera is attached to a new satellite, and this process continues until all satellites are visited. Once all satellites in the batch are visited, a new batch of 1,000 satellites is introduced, and the image capture process repeats until the dataset reaches the desired size. The simulation automatically annotates each image by generating and recording bounding boxes around all satellites visible within a 5 km range of the camera.

To categorize the images according to distance, we classify each image into one of three distance ranges: 0 to 0.5 km, 0.5 to 2 km, and 2 to 5 km. For every captured image, we identify the closest satellite using the label metadata, and the image is assigned to a category based on this distance. To balance the dataset, we ensure an equal number of images per category.

\section{Performance Evaluation}

\subsection{Experimental Setup}
To evaluate our models on SODv2, we trained each model on 450 images and tested it on a separate set of 150 images. For each training run, the epoch that achieved the best mAP50 was recorded.

We conducted our experiments on a high-performance computing cluster with a Tesla V100-32GB GPU, logging memory usage and inference time for each run. Additionally, we utilized a Jetson Orin Nano to simulate onboard conditions of LEO satellites, focusing on real-time inference in a resource-constrained environment.

To ensure reproducibility, each model was trained using three fixed values for random seed: seed = 1, seed = 2, and seed = 3. The seed ensured consistency in weight initialization, data shuffling, and augmentation. The YOLOv9 framework was used to set the seed in all relevant libraries, ensuring the repeatability of the experiment. Specifically, YOLOv9 initialized the seed for Python’s built-in random module, NumPy, and PyTorch. 

For computational efficiency measurements on the V100 GPU, we averaged results over 25 consecutive runs, discarding the first five to minimize the impact of resource allocation variability. We used the \texttt{seff} command to monitor memory usage and the \texttt{thop} library to compute GFLOPs and model parameters.

On the Jetson Orin Nano, we used the \texttt{tegrastats} tool to track memory usage while monitoring CPU and GPU power consumption during inference. Baseline measurements were recorded before inference and subtracted from logged power and memory usage values. We then computed both the mean and peak values for each run and averaged them across all runs to obtain final results.

All models were trained for 1000 epochs with a batch size of 16. The hyperparameter details were provided in \cite{zhang2024sensing}. To accurately measure the real-world inference performance of our models, we set the batch size to 1 during inference testing. These settings remained fixed across all experiments to allow fair comparisons. Upon completing each training run, we evaluated model performance by measuring mAP50 and mAP50:95 on a test set of 150 images.

\begin{table}[ht]
    \centering
    \caption{Evaluation Results for 0 km to 5 km Distance Range.}
    
    \begin{tabular}{lccccc}
        \toprule
        Model & Random Seed & mAP50 & mAP50:95 & GFLOPs & parameter count  \\
        \midrule
        \multirow{4}{*}{GELAN-t} 
            & 1   & 0.738 & 0.277 & \multirow{4}{*}{7.3} & \multirow{4}{*}{1913443} \\
            & 2   & 0.717 & 0.264 \\
            & 3   & 0.707 & 0.256 \\
            & Avg. & 0.721 & 0.266\\
        \midrule
        \multirow{4}{*}{GELAN-SE} 
            & 1   & 0.739 & 0.272 & \multirow{4}{*}{5.3} & \multirow{4}{*}{1301011} \\
            & 2   & 0.724 & 0.265 \\
            & 3   & 0.722 & 0.274 \\
            & Avg. & 0.728 & 0.270 \\
        \midrule
        \multirow{4}{*}{GELAN-ViT} 
            & 1   & 0.739 & 0.270 & \multirow{4}{*}{5.6} & \multirow{4}{*}{7345068} \\
            & 2   & 0.710 & 0.252 \\
            & 3   & 0.763 & 0.273 \\
            & Avg. & 0.737 & 0.265 \\
        \midrule
        \multirow{4}{*}{GELAN-ViT-SE} 
            & 1   & 0.761 & 0.280 & \multirow{4}{*}{5.6} & \multirow{4}{*}{7362012} \\
            & 2   & 0.757 & 0.278 \\
            & 3   & 0.736 & 0.265 \\
            & Avg. &  \textbf{0.751} &  \textbf{0.274} \\
        \midrule
        \multirow{4}{*}{GELAN-RepViT} 
            & 1   & 0.728 & 0.259 & \multirow{4}{*}{5.2} & \multirow{4}{*}{1264459} \\
            & 2   & 0.719 & 0.268 \\
            & 3   & 0.723 & 0.271 \\
            & Avg. & 0.723 & 0.266 \\
        \midrule
        \multirow{4}{*}{GELAN-RepViT-SE} 
            & 1   & 0.739 & 0.271 & \multirow{4}{*}{5.2} & \multirow{4}{*}{1275107} \\
            & 2   & 0.735 & 0.264 \\
            & 3   & 0.750 & 0.262 \\
            & Avg. & 0.741 & 0.266 \\
        \bottomrule
    \end{tabular}
    \label{tab:overview}
\end{table}

Table \ref{tab:overview} compares the performance of the models on the SODv2 dataset. GELAN-t \cite{wang2024yolov9learningwantlearn} is used as the baseline model for comparison. The proposed models are evaluated in two comparison pairs: GELAN-ViT vs. GELAN-ViT-SE and GELAN-RepViT vs. GELAN-RepViT-SE. Additionally, although GELAN-SE has a different architecture than GELAN-t, it forms a pair with GELAN-t to evaluate its performance as a purely CNN-based model. Furthermore, comparing GELAN-SE with GELAN-ViT-SE and GELAN-RepViT-SE allows us to isolate architectural differences and analyze the impact of ViT integration.

\subsection{Performance Results on V100 GPU}

Table \ref{tab:overview} shows that the addition of an SE block increases mAP50 in the three pairs of models. For GELAN-t, the mAP50 increases from 0.721 to 0.728 with SE. GELAN-RepViT exhibits an improvement of 0.723 to 0.741, while GELAN-ViT increases from 0.737 to 0.751. Among all models tested, GELAN-ViT-SE shows the highest overall mAP50 of 0.751. Analyzing mAP50:95, we observe improvements in most cases, though the impact of SE differs across models. GELAN-SE shows an increase from 0.266 to 0.270, while GELAN-ViT-SE shows a rise from 0.265 to 0.274, marking the highest final mAP50:95. However, GELAN-RepViT maintains an mAP50:95 of 0.266 with SE, indicating no improvement from its baseline. Overall, these results suggest that the SE block generally enhances detection performance across different architectures, with the most significant improvements in ViT-based models, particularly in mAP50.

Table \ref{tab:overview} also includes the results for each random seed used during training, showing the performance range of each model. GELAN-t's mAP50 ranges from 0.707 to 0.738 with a spread of 0.031. GELAN-SE achieves a narrower range of 0.722 to 0.739, reducing its spread to 0.017. GELAN-RepViT achieves a range of 0.719 to 0.728, maintaining a narrower spread of 0.009. GELAN-RepViT-SE achieves performance of 0.735 to 0.750, slightly increasing the range to 0.015. GELAN-ViT has a range of 0.710 to 0.763, with a spread of 0.053, the largest among all models. GELAN-ViT-SE narrows the range to 0.736 to 0.761, reducing the spread to 0.025, while also producing the highest overall mAP50 score of 0.751.

Regarding mAP50:95, GELAN-t has a range of 0.256 to 0.277, with a spread of 0.021. GELAN-SE achieves a slightly higher and narrower range of 0.265 to 0.274, reducing its spread to 0.009. GELAN-RepViT operates within 0.259 to 0.271, with a spread of 0.012. GELAN-RepViT-SE achieves a range of 0.262 to 0.271, slightly narrowing the spread to 0.009, while consistently achieving higher scores. GELAN-ViT achieves a range between 0.252 and 0.273, with a spread of 0.021, similar to GELAN-t. However, GELAN-ViT-SE exhibits a narrower range of 0.265 to 0.280, with a spread of 0.015, while also providing the highest overall mAP50:95 score of 0.280.

\begin{table}[ht]
\centering
\caption{Comparison of Inference Time and Memory Usage for different models on V100 GPU}
\begin{tabular}{|l|c|c|}
\hline
\textbf{Model} & \textbf{Average Inference Time (ms)} &  \textbf{Memory Usage (GB)} \\ \hline
GELAN-t & 2.34 & 3.51 \\ \hline
GELAN-SE & 2.17 & 3.23 \\ \hline
GELAN-ViT &  2.59 & 3.29 \\ \hline
GELAN-ViT-SE & 2.71 & 3.65 \\ \hline
GELAN-RepViT & 2.13 & 3.35 \\ \hline
GELAN-RepViT-SE  & 2.33  & 3.53 \\ \hline
\end{tabular}
\label{tab:inference_v100}
\end{table}

\subsection{Inference Result on V100 GPU} 

Table \ref{tab:inference_v100} shows the inference time and memory usage for each model when tested on an NVIDIA V100 GPU. Among all models, GELAN-RepViT shows the fastest inference time at 2.13 ms, while adding the SE block increases it slightly to 2.33 ms. GELAN-ViT achieves a slower inference time of 2.59 ms, which extends to 2.71 ms in GELAN-ViT-SE. Meanwhile, GELAN-t, serving as the baseline model, shows an inference time of 2.34 ms, while GELAN-SE has a slightly faster time of 2.17 ms, due to its small size in GFLOPs.

Memory consumption is also recorded in Table \ref{tab:inference_v100}. GELAN-t requires 3.51 GB of memory, while GELAN-SE uses 3.23 GB. GELAN-ViT and GELAN-ViT-SE show an increase in memory usage, requiring 3.29 GB and 3.65 GB, respectively. Similarly, GELAN-RepViT consumes 3.35 GB, while GELAN-RepViT-SE increases memory usage slightly to 3.53 GB.

\subsection{Result Discussion on V100 GPU}

The consistent gains from the integration of the SE block in all variants of GELAN underscore the importance of channel-wise recalibration to refine local characteristics. Although GELAN-ViT and GELAN-RepViT already allocate a dedicated capacity $C(F_l)$ for local features, treating every channel uniformly can limit the model’s ability to emphasize the most relevant features. By assigning higher attention scores to the most discriminative channels, SE effectively enhances feature extractions by emphasizing important details.

A comparison of mAP50 ranges shows that GELAN-ViT has the highest variation (0.053), although its overall scores remain higher than GELAN-t. For mAP50:95, GELAN-ViT-SE consistently achieves the highest performance while narrowing its spread from 0.021 to 0.015. Although GELAN-t and GELAN-SE exhibit smaller variations in mAP50 scores, their overall performance remains lower than that of ViT-based models. Given the minimal increase in parameters and GFLOPs, the overall trade-off appears advantageous, reinforcing the effectiveness of the SE block. 
The SE block also causes a slight increase in both inference time and memory usage. However, this increase is small and does not significantly affect real-time deployment.

\subsection{Limitations and Multi-Object Detection Performance}

\begin{table}[ht]
    \centering
    \caption{Evaluation Results on Multi-Object Detection.}
\resizebox{1\textwidth}{!}{%
    \begin{tabular}{lcccccc}
        \toprule
        Model & Seed & mAP50 (Adjusted) & mAP50:95 (Adjusted) & mAP50 (Default) & mAP50:95 (Default)  \\
        \midrule
        \multirow{4}{*}{GELAN-t} 
            & 1   & 0.723 & 0.298 & 0.701 & 0.271 \\
            & 2   & 0.729 & 0.291 & 0.694 & 0.275 \\
            & 3   & 0.738 & 0.309 & 0.715 & 0.285 \\
            & Avg. & \textbf{0.730} & \textbf{0.299} & 0.703 & 0.277 \\
        \midrule
        \multirow{4}{*}{GELAN-SE} 
            & 1   & 0.731 & 0.296 & 0.703 & 0.273 \\
            & 2   & 0.728 & 0.309 & 0.707 & 0.279 \\
            & 3   & 0.730 & 0.291 & 0.703 & 0.277 \\
            & Avg. & \textbf{0.730} & \textbf{0.299} & 0.704 & 0.276 \\
        \midrule
        \multirow{4}{*}{GELAN-ViT} 
            & 1   & 0.727 & 0.301 & 0.709 & 0.283 \\
            & 2   & 0.727 & 0.295 & 0.703 & 0.275 \\
            & 3   & 0.723 & 0.289 & 0.701 & 0.276 \\
            & Avg. & 0.726 & 0.295 & 0.704 & 0.278 \\
        \midrule
        \multirow{4}{*}{GELAN-ViT-SE} 
            & 1   & 0.734 & 0.290 & 0.725 & 0.289 \\
            & 2   & 0.724 & 0.294 & 0.680 & 0.255 \\
            & 3   & 0.719 & 0.291 & 0.695 & 0.267 \\
            & Avg. & 0.726 & 0.292 & 0.700 & 0.270 \\
        \midrule
        \multirow{4}{*}{GELAN-RepViT} 
            & 1   & 0.725 & 0.298 & 0.705 & 0.283 \\
            & 2   & 0.722 & 0.284 & 0.722 & 0.286 \\
            & 3   & 0.724 & 0.292 & 0.707 & 0.288 \\
            & Avg. & 0.724 & 0.291 & 0.711 & 0.286 \\
        \midrule
        \multirow{4}{*}{GELAN-RepViT-SE} 
            & 1   & 0.716 & 0.286 & 0.716 & 0.286 \\
            & 2   & 0.716 & 0.288 & 0.716 & 0.288 \\
            & 3   & 0.734 & 0.304 & 0.707 & 0.282 \\
            & Avg. & 0.722 & 0.293 & \textbf{0.713} & \textbf{0.285} \\
        \bottomrule
    \end{tabular}
    }
    \label{tab:multi_object}
\end{table}

While our proposed method performs well on certain datasets, it does not consistently outperform GELAN-t in all tasks due to the reduction in network parameters in GELAN-s-Reduced, which limits effectiveness in some scenarios. Specifically, we evaluate our model on SODv2 but with multi-object detection, where each image contains 9-18 satellite objects. Each model is trained using the same random seeds as in Table \ref{tab:overview}, and we focus on the average performance across these runs to ensure consistency in evaluation. The performance is summarized in Table \ref{tab:multi_object}, which compares multi-object detection results under both adjusted and default hyperparameters. The adjusted hyperparameters refer to the tuned configuration from our previous work \cite{zhang2024sensing}, while the default hyperparameters are those provided by YOLOv9.

Using adjusted hyperparameters, GELAN-t achieves an mAP50 of 0.730 and an mAP50:95 of 0.299, while GELAN-SE achieves the same mAP50 and mAP50:95 of 0.730 and 0.299. GELAN-ViT obtains an mAP50 of 0.726 and an mAP50:95 of 0.295, and GELAN-ViT-SE reaches an mAP50 of 0.726 and an mAP50:95 of 0.292. GELAN-RepViT shows an mAP50 of 0.724 and an mAP50:95 of 0.291, with GELAN-RepViT-SE showing 0.722 and 0.293, respectively.

GELAN-t and GELAN-SE achieve the highest mAP50 of 0.730, while GELAN-SE achieves the highest mAP50:95 of 0.299. GELAN-RepViT-SE achieves the lowest mAP50 of 0.722, while GELAN-RepViT has the lowest mAP50:95 of 0.291. All models show similar average performance, differing by less than 0.008 in mAP50 and 0.009 in mAP50:95.

To assess robustness across hyperparameter settings, we re-evaluate the models using YOLOv9’s default hyperparameters. Under these conditions, GELAN-ViT achieves an mAP50 of 0.704 and an mAP50:95 of 0.278 compared to GELAN-t's 0.703 and 0.277. GELAN-SE receives an mAP50 of 0.704 and an mAP50:95 of 0.276. GELAN-ViT-SE achieves an mAP50 of 0.700 and an mAP50:95 of 0.270. GELAN-RepViT achieves an mAP50 of 0.711 and an mAP50:95 of 0.286, while GELAN-RepViT-SE shows an mAP50 of 0.713 and an mAP50:95 of 0.285.

With the default hyperparameters, GELAN-RepViT-SE has the highest mAP50 of 0.713, while GELAN-RepViT has the highest mAP50:95 performance of 0.286. Although GELAN-ViT-SE achieves the lowest mAP50 and mAP50:95 of 0.700 and 0.270, it also achieves the highest single-run mAP50 and mAP50:95 of 0.725 and 0.289. All models show similar average performance, differing in mAP50 by less than 0.013 and in mAP50:95 by 0.016.

Given these findings, we cannot definitively conclude that one model is superior to the other in multi-object detection tasks. Instead, these results suggest that performance variations depend on dataset and hyperparameter settings rather than architectural differences alone.

\begin{table}[ht]
\centering

\caption{Summary of inference time, memory and power usage for each model tested on Jetson Orin Nano}
\begin{tabular}{|l|c|c|c|c|c|c|c|}
\hline
\textbf{Model Name} & \textbf{Inference Time} & \textbf{Peak RAM} & \textbf{Average RAM}  & \textbf{Peak Power} & \textbf{Average Power} \\
\hline
GELAN-t & 46.14 ms & 2.377 GB & 2.376 GB & 2080.7 mW & 1829.6 mW\\
\hline
GELAN-SE & 39.53 ms & 2.407 GB & 2.402 GB & 2023.5 mW & 1799.2 mW\\
\hline
GELAN-ViT & 56.47 ms & 2.463 GB & 2.461 GB & 1988.6 mW & 1793.2 mW\\
\hline
GELAN-ViT-SE & 58.88 ms & 2.557 GB & 2.556 GB & 2028.7 mW & 1799.9 mW\\
\hline
GELAN-RepViT & 37.40 ms & 2.395 GB & 2.378 GB & 1986.2 mW & 1756.1 mW\\
\hline
GELAN-RepViT-SE & 39.89 ms & 2.401 GB & 2.380 GB & 2028.4 mW & 1788.0 mW\\
\hline
\end{tabular}%
\label{tab:memory_power_usage}
\end{table}

\subsection{Evaluation on NVIDIA Jetson Orin Nano}
\begin{figure}[H]
    \centering
    \includegraphics[width=1\textwidth]{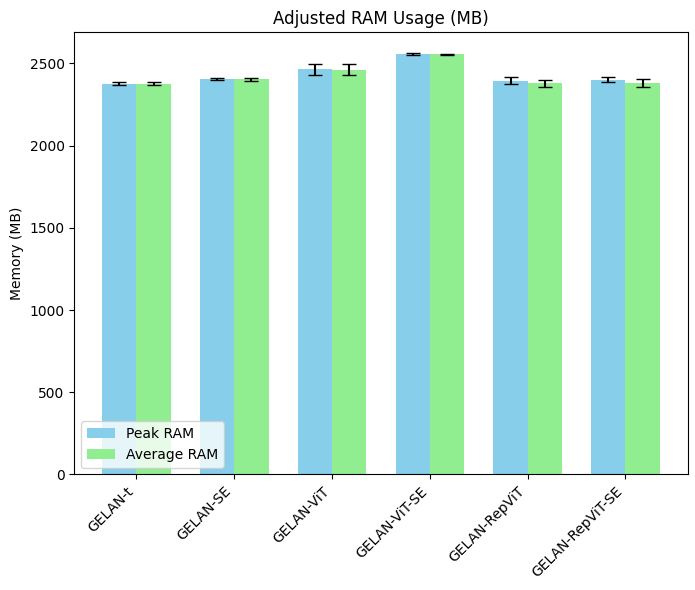}
    \caption{Bar charts of RAM usage on Jetson Orin Nano, with the 95\% confidence level indicated for each model.}
    \label{fig:memory_chart}
\end{figure}
\begin{figure}[ht]
    \centering
    \includegraphics[width=1\textwidth]{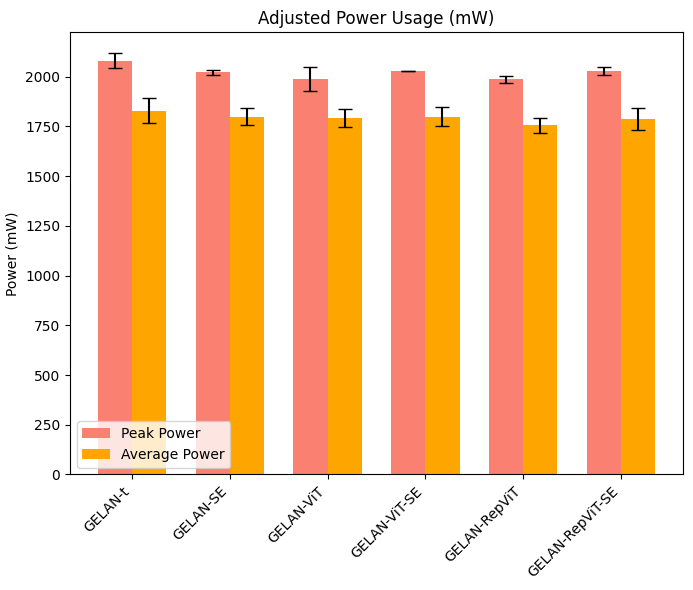}
    \caption{Bar charts of power usage on Jetson Orin Nano, with the 95\% confidence level indicated for each model.}
    \label{fig:power_chart}
\end{figure}
While the inference time differences are small on a high-performance GPU (0.58 ms difference between the fastest and slowest models), they have a greater impact in embedded environments where computational resources are limited. To further assess the efficiency of these models under real-world constraints, we evaluated inference time, memory consumption, and power usage on an NVIDIA Jetson Orin Nano.

Table \ref{tab:memory_power_usage} provides a comparison of the models on the NVIDIA Jetson Orin Nano. The results are visualized in Fig. \ref{fig:memory_chart} and Fig. \ref{fig:power_chart}, which also indicate the 95\% confidence intervals for each model. The table includes key metrics such as inference time, memory usage, and power consumption. 

GELAN-RepViT achieves the fastest inference time at 37.40 ms, followed by GELAN-RepViT-SE at 39.89 ms and GELAN-SE at 39.53 ms. In contrast, GELAN-t requires 46.14 ms, while ViT-based models exhibit slower inference times, with GELAN-ViT at 56.47 ms and GELAN-ViT-SE at 58.88 ms.

The peak and average RAM usage follow a similar pattern across all models. GELAN-t peaks at 2.377 GB with an average of 2.376 GB, whereas GELAN-SE exhibits slightly higher values at 2.407 GB and 2.402 GB, respectively. The ViT-based models show higher memory consumption, with GELAN-ViT peaking at 2.463 GB (2.461 GB average) and GELAN-ViT-SE reaching the highest usage at 2.557 GB (2.556 GB average). In contrast, GELAN-RepViT and GELAN-RepViT-SE maintain lower memory usage, with peak values of 2.395 GB and 2.401 GB and average values of 2.378 GB and 2.380 GB, respectively.

In relative terms, GELAN-SE shows a 1.26\% increase in peak RAM and a 1.09\% increase in average RAM over GELAN-t. GELAN-ViT exhibits a 3.62\% peak RAM increase and a 3.58\% average RAM increase. GELAN-ViT-SE has the highest increase, with 7.57\% more peak RAM and 7.58\% more average RAM. Meanwhile, GELAN-RepViT remains the most memory-efficient, with only a 0.76\% peak RAM increase and a 0.08\% average RAM increase. GELAN-RepViT-SE exhibits a 1.01\% peak RAM increase and a 0.17\% average RAM increase.

Regarding power consumption, GELAN-t shows the highest peak power of 2080.7 mW, while the other models show values ranging from approximately 1896 mW to 2028.7 mW, representing reductions of 4.54\% to 2.50\% compared to GELAN-t. In terms of average power, GELAN-t consumes 1829.6 mW, compared to GELAN-RepViT's lowest average of 1756.1 mW (4.02\% lower), while the remaining models are tightly clustered between 1788 mW and 1799 mW (reductions of 1.62\% to 2.27\%).

\subsection{Result Discussion on Jetson Orin Nano}

The integration of the SE block has a small impact on performance metrics. For example, adding the SE block to GELAN-RepViT increases the inference time from 37.40 to 39.89 ms, an increase of approximately 2.49 ms. Similarly, in the ViT-based branch, the SE block results in a small increase of around 2.41 ms, from 56.47 ms to 58.88 ms. 

In terms of memory usage, GELAN-ViT shows an increase in both peak and average RAM usage compared to the baseline GELAN-t, with GELAN-ViT-SE showing a further increase of approximately 0.094–0.095 GB over GELAN-ViT. In contrast, the RepViT-based variants remain largely consistent in their memory usage. 

In terms of power, the integration of the SE block results in only minimal differences across models, with all proposed variants exhibiting slightly lower peak and average power consumption compared to the baseline GELAN-t. Among them, RepViT-based models remain the most efficient overall.

Although the SE block introduces minor increases in inference time and power consumption, the trade-off appears justifiable given the overall improvements in detection accuracy. The power consumption differences between models are minimal and do not confer a clear advantage or disadvantage in resource efficiency. However, GELAN-RepViT remains the most power-efficient option, making it a strong candidate for low-power embedded environments.

While GELAN-ViT-SE generally consumes more memory and power, it offers improved detection performance in SOD scenarios, such as detecting small or partially occluded objects. However, given that onboard SOD tasks are typically resource-constrained, GELAN-RepViT-SE provides a more practical balance between efficiency and accuracy.

\FloatBarrier
\section{Conclusion}
In this paper, we introduced the SODv2 dataset and proposed three new models, GELAN-SE, GELAN-ViT-SE, and GELAN-RepViT-SE, to evaluate the impact of integrating SE blocks on the GELAN architecture. Our experiments showed that the addition of the SE block improved detection accuracy and stability while maintaining similar computational efficiency, which confirmed that channel-wise recalibration in the GELAN framework is a promising strategy for onboard SOD tasks. 

Overall, these results underscore the effectiveness of integrating the SE block into GELAN-based architectures to achieve robust and efficient performance in the SOD task. A future direction would be to expand the scope of experiments by testing on diverse datasets, evaluating additional random seeds, and exploring different hyperparameter configurations to analyze the advantages of each model.

\section*{Acknowledgment}
We acknowledge the support of the Natural Sciences and Engineering Research Council of Canada (NSERC), [funding reference number RGPIN-2022-03364].

\bibliography{references}
\end{document}